\documentclass{article}
\usepackage{ijcai26}

\usepackage{times}
\usepackage{soul}
\usepackage{url}
\usepackage[hidelinks]{hyperref}
\usepackage[utf8]{inputenc}
\usepackage[small]{caption}
\usepackage{graphicx}
\usepackage{amsmath}
\usepackage{amsthm}
\usepackage{amssymb} 
\usepackage{booktabs}
\usepackage{algorithm}
\usepackage{algorithmic}
\usepackage[switch]{lineno}
\DeclareMathOperator*{\argmax}{arg\,max}

\linenumbers

\title{\textsc{BoolXllm}: LLM-Assisted Explainability for Boolean Models}

\author{
Du Cheng\textsuperscript{\rm 1},
Serdar~Kad{\i}o\u{g}lu\textsuperscript{\rm 1,2},
    Xin Wang\textsuperscript{\rm 1}
\affiliations
     \textsuperscript{\rm 1} AI Center of Excellence, Fidelity Investments, MA, USA \\
    \textsuperscript{\rm 2} Department of Computer Science, Brown University, RI, USA\\
\emails
serdark@cs.brown.edu
}

\usepackage{pythonhighlight}
\usepackage{booktabs}

\usepackage{bm}
\usepackage{amsmath}
\usepackage{rotating}
\usepackage{makecell}

\begin{document}
\nolinenumbers
\maketitle

\begin{abstract}
Interpretable machine learning aims to provide transparent models whose decision-making processes can be readily understood by humans. Recent advances in rule-based approaches, such as expressive Boolean formulas (BoolXAI), offer faithful and compact representations of model behavior. However, for non-technical stakeholders, main challenges remain in practice: (i) selecting semantically meaningful features and (ii) translating formal logical rules into accessible explanations.

In this work, we propose \textsc{BoolXllm} 
, as a hybrid framework that integrates Large Language Models (LLMs) into the end-to-end pipeline of Boolean rule learning. We augment \textsc{BoolXai} , an expressive Boolean rule-based classifier, with LLMs at three critical stages: (1) feature selection, where LLMs guide the identification of domain-relevant variables; (2) threshold recommendation, where LLMs propose semantically meaningful discretization strategies for numerical features; and (3) rule  compression and interpretation, where Boolean rules are translated into natural language explanations at both global and local levels.

This integration bridges formal, faithful explanations with human-understandable narratives. This allows build an explainable AI system that is both theoretically grounded and accessible to non-experts. Early empirical results demonstrate that LLM-assisted pipelines improve interpretability while maintaining competitive predictive performance. Our work highlights the promise of combining symbolic reasoning with language-based models for human-centered explainability.

\end{abstract}

\section{Introduction}
\label{sec:introduction}

As machine learning (ML) systems continue to be deployed in high-stakes domains such as finance, healthcare, and policymaking, the need for eXplainable AI (XAI) has become increasingly critical. Interpretable machine learning models, which provide transparent and self-explanatory representations of their decision processes, have emerged as a compelling alternative to post-hoc explanation methods. 

Among these, rule-based models offer a natural and intuitive form of explanation, particularly when expressed as logical formulas. Recent work has introduced expressive Boolean rule-based approaches, such as \textsc{BoolXai}~\cite{KadiogluZRBSSZK25,make5040086}\footnote{\url{https://github.com/fidelity/boolxai}}, which extend classical logical representations with operators such as \texttt{AtLeast}, \texttt{AtMost}, and \texttt{Choose}. These models strike a balance between fidelity and interpretability by learning compact logical rules that accurately capture patterns in the data. Moreover, such approaches have demonstrated practical viability in real-world applications, including enterprise-scale deployments and explainability-as-a-service platforms. 

Despite these advances, two important challenges persist in operational settings. First, when dealing with high-dimensional datasets, practitioners must decide which features to include in the model and how to discretize numerical variables into meaningful categories. These decisions are often guided by domain expertise and can significantly affect both model performance and interpretability. Second, although Boolean rules are structurally simple, they may still appear abstract or difficult to interpret for non-technical stakeholders. For example, a rule such as \texttt{Or(duration > 393, nr.employed <= 5076, month = mar)} is concise, yet lacks semantic grounding without additional context.

In this paper, we address these challenges by proposing a hybrid explainability framework that integrates LLMs into the life-cycle of interpretable model development. Our approach, termed \textsc{BoolXllm}, augments Boolean rule learning with language-based reasoning to enhance usability and accessibility. Specifically, we incorporate LLMs at three stages of the pipeline:

\begin{enumerate}
    \item \textbf{Feature Selection}, where LLMs guide the identification of semantically meaningful and domain-relevant inputs 
    \item \textbf{Threshold Recommendation}, where LLMs propose context-aware discretization thresholds for numerical features, and 
    \item \textbf{Rule Compression \& Interpretation}, where learned Boolean rules are translated into natural language explanations at both global (population-level) and local (instance-level) perspectives.

\end{enumerate}

Our framework, \textsc{BoolXllm}, combines the strengths of symbolic and statistical paradigms: the logical rigor and faithfulness of Boolean rules with the linguistic expressiveness and contextual awareness of LLMs. By grounding LLM-generated explanations in formally defined rules, we mitigate concerns related to hallucination and preserve fidelity to the underlying model. At the same time, the integration of LLMs enables explanations that are more accessible and aligned with human reasoning.

We evaluate our approach on benchmark datasets and real-world scenarios, demonstrating that LLM-assisted pipelines in \textsc{BoolXllm} can improve interpretability and usability while maintaining competitive predictive performance. Our results suggest that combining interpretable models with language-based interfaces provides a promising direction for advancing human-centered XAI.

In summary, the contributions of this paper are as follows:
\begin{itemize}
    \item We introduce a novel framework for \emph{LLM-assisted explainability} that integrates Large Language Models into interpretable rule-based learning via \textsc{BoolXllm}.
    \item We propose methods for LLM-guided feature selection and threshold recommendation to improve the semantic grounding of Boolean rules.
    \item We develop global and local rule interpretation techniques using LLMs to translate formal logic into human-understandable explanations.
    \item We empirically demonstrate the effectiveness of our approach in improving interpretability while preserving predictive performance.
\end{itemize}

\section{Background: \textsc{BoolXai}}
\label{sec:boolxai}
Let us start with a brief overview of \textsc{BoolXai}~\cite{KadiogluZRBSSZK25}, an interpretable machine learning approach based on expressive Boolean formulas. This serves as the foundation for our proposed \textsc{BoolXllm} framework.

\subsection{Problem Formulation}
We consider supervised binary classification with a dataset $(X, y)$, where $X \in \mathbb{R}^{n \times d}$ denotes the feature matrix and $y \in \{0,1\}^n$ the corresponding labels. The goal of \textsc{BoolXai} is to learn a Boolean rule $R$ that maps input features to predictions while balancing predictive performance and interpretability.

This is formulated as a \emph{Rule Optimization Problem (ROP)}:
\begin{equation}
R^* = \argmax_{R} \left[ S(R(X), y) - \lambda C(R) \right] \quad \text{s.t.} \quad C(R) \leq C',
\end{equation}
where $S$ is a performance metric (e.g., balanced accuracy), $C(R)$ measures the complexity of the rule, $C'$ is a user-defined complexity bound, and $\lambda$ controls the trade-off between performance and interpretability.

The central idea is to learn rules that are both accurate and concise, as increasing complexity generally reduces interpretability.

\subsection{Expressive Boolean Rules}

\textsc{BoolXai} represents models as Boolean formulas over input features. The main novelty of \textsc{BoolXai} is that, in addition to classical logical operators such as \texttt{And} and \texttt{Or}, it introduces expressive, parameterized operators such as \texttt{AtLeast}$(k, \cdot)$, \texttt{AtMost}$(k, \cdot)$, \texttt{Choose}$(k, \cdot)$


These operators enable compact and semantically meaningful representations of decision logic that would otherwise require significantly more complex structures (e.g., deep decision trees or large conjunctive normal forms). 

As an example, consider a simple scenario with five binary features that lead to a positive label only if at least three of the features hold. In \textsc{BoolXai}, this is straightforward to state as {\texttt{AtLeast3($f_0, \dots, f_4$)}}. Decision Trees are generally considered highly interpretable, but they fail in this case. Training a decision tree for this simple case leads to a surprisingly large tree with \textit{19 split nodes}. This decision tree is deep and difficult to interpret. If a user faces it, it is hard to comprehend the global view of \texttt{AtLeast3} that the tree is trying to enforce. Alternatively, instead of our expressive operators, if we only consider propositional Boolean formulas given Conjunctive Normal Form (CNF) using disjunctions and conjunctions, then we need a CNF over \textit{13 clauses} using \textit{11 variables} and a total of \textit{29 literals} across the clauses. In comparison, expressive Boolean formula offers a concise representation with a single \texttt{AtLeast}.
\textsc{BoolXai} can be visualized as syntax trees whose nodes correspond to operators and leaves correspond to literals (features or their negations).

\subsection{Optimization via Local Search}

To learn Boolean rules efficiently, \textsc{BoolXai} employs a native local search strategy that operates directly in the space of valid formulas. Starting from a randomly initialized rule, the algorithm iteratively applies local modifications, including adding or removing literals, swapping operators, and restructuring subtrees~\cite{make5040086}.

The optimization is guided by a simulated annealing scheme, which balances exploration and exploitation by probabilistically accepting candidate moves based on the change in the objective function. This approach enables scalable search over large combinatorial spaces without requiring specialized solvers such as integer linear programming or MaxSAT. Moreover, it is shown to be quantum amenable. 

\subsection{Practical Considerations and Limitations}

\textsc{BoolXai} demonstrated strong performance and usability in practice, including deployment as an explainability-as-a-service (EaSe) in enterprise settings~\cite{KadiogluZRBSSZK25}. Its design prioritizes scalability, interoperability (e.g., with scikit-learn pipelines), and ease of use through high-level abstractions and visualization tools.

Despite these strengths, several challenges remain in operational settings. First, practitioners must manually select relevant features from potentially high-dimensional datasets, which requires significant domain expertise. Second, numerical features must be discretized into thresholded conditions, where arbitrary or poorly chosen thresholds can hinder interpretability. Finally, although Boolean rules are structurally concise, they may still appear abstract to non-technical users, limiting their accessibility and practical impact.

These limitations motivate the integration of language-based models, as explored in our  \textsc{BoolXllm} framework.

\begin{figure*}[t]
\includegraphics[width=\textwidth]{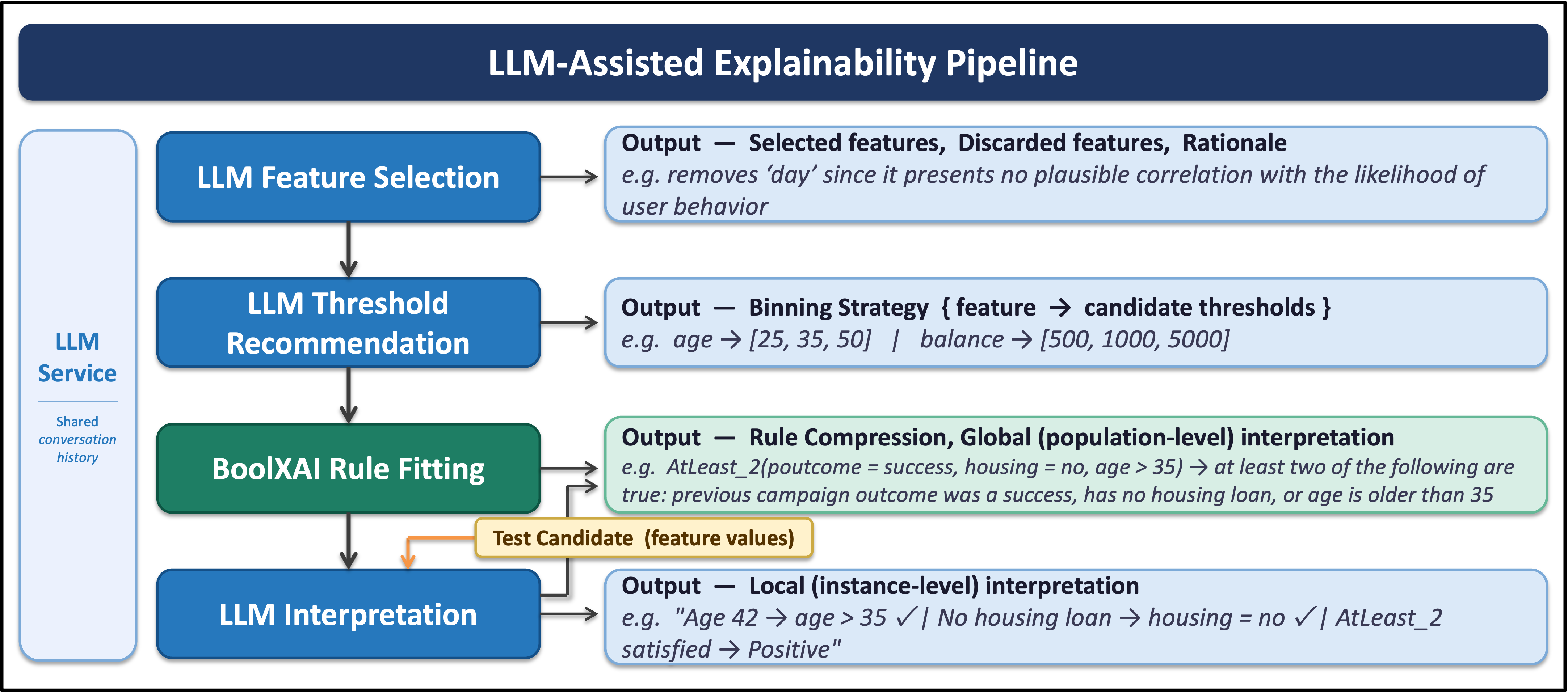}
\caption{\textbf{\textsc{BoolXlmm}} architecture highlighting three stages where LLMs are incorporated to enhance explainability: (1) LLM Feature Selection, which identifies semantically meaningful and business-relevant features; (2) LLM Threshold Recommendation, which proposes context-aware discretization thresholds for numerical variables to improve semantic clarity; (3) LLM-Assisted Rule Compression and Interpretation, which produces human-readable explanations at both global and local levels.
}
\label{fig:boolxlmm}
\vspace{-0.4cm}
\end{figure*}

\section{\textsc{BoolXllm}: LLM-Assisted Explainability}
\label{sec:boolxllm}

As illustrated in Figure~\ref{fig:boolxlmm}, we propose a novel integration of LLMs into the \textsc{BoolXai} pipeline. We enable \emph{LLM-assisted explainability} at three key stages:

\begin{itemize}
    \item \textbf{Feature Selection:} LLMs guide the identification of semantically meaningful and business-relevant features, reducing the dimensionality of the search space while preserving interpretability.
    
    \item \textbf{Threshold Recommendation:} For numerical variables, LLMs propose context-aware discretization thresholds that align with domain semantics (e.g., meaningful ranges in finance or marketing), improving the interpretability of resulting rules.
    
    \item \textbf{Rule Interpretation:} LLMs translate Boolean rules into human-readable explanations at both:
    \begin{itemize}
        \item \emph{Global level:} summarizing population-level patterns captured by the rule.
        \item \emph{Local level:} explaining individual predictions in natural language.
    \end{itemize}
\end{itemize}

This hybrid framework combines the formal guarantees of symbolic, rule-based models with the linguistic expressiveness of LLMs. Notice that LLM explanations are grounded in the underlying Boolean rule to ensure fidelity and avoid hallucinated reasoning. Our goal is to enable an explainable AI system that is both faithful and accessible to stakeholders. 

\subsection{Feature Selection}
Selecting relevant features is critical for both the performance and interpretability of Boolean rules. In high-dimensional settings, this process typically depends on domain expertise. We address this challenge by integrating LLMs as a semantic interface between dataset metadata and feature selection.

\smallskip
\noindent\textbf{Prompt Design:} We formulate feature selection as a structured prompting task in which the LLM receives (i) a natural language prediction objective, (ii) dataset metadata (feature names and descriptions), and (iii) optional domain context. The model is instructed to select and rank features based on their semantic relevance to the task and to return a constrained JSON output, enabling deterministic parsing and validation. The full prompt template is provided in Appendix~\ref{appendix:feature_selection_prompt}.

\smallskip
\noindent\textbf{Domain Knowledge:} A key advantage of this approach is the incorporation of implicit domain knowledge. Unlike purely statistical methods, LLMs leverage prior understanding of common business relationships to prioritize conceptually meaningful variables (e.g., engagement-related features for marketing), even when empirical correlations are weak.

\smallskip
\noindent\textbf{Semantic Grounding via Dataset Metadata:} We emphasize semantic grounding by anchoring LLMs with dataset metadata. Providing feature names and descriptions enables interpretation of variables as real-world concepts rather than abstract tokens, while constraining outputs to the given schema mitigates hallucination.

\subsection{Threshold Recommendation}
Boolean rule learning requires discretizing numerical features into threshold conditions, which directly impacts interpretability. Standard binning methods often produce arbitrary splits (e.g., \texttt{age $>$ 36.33)}, which lack meaningful differentiation. We address this limitation by using LLMs to generate domain-aware, interpretable thresholds.
We formulate threshold selection as a structured prompt where LLMs receive feature metadata (names, descriptions, units) and the prediction objective, and is instructed to propose thresholds that map continuous variables into reasonable bins. The output is constrained to a structured JSON format for deterministic parsing. The prompt template is provided in Appendix~\ref{appendix:threshold_prompt}.

The resulting thresholds are used to discretize numerical features prior to Boolean rule learning. This yields conditions aligned with human reasoning (e.g., \texttt{age $>$ 35, balance $>$ 500)} rather than arbitrary numeric boundaries. Multiple candidate thresholds can be generated and evaluated within the \textsc{BoolXai} optimization process.

\subsection{Rule Compression and Interpretation}
Although Boolean rules learned by \textsc{BoolXai} are compact, arbitrarily different rules might achieve similar classification performance which makes it difficult for stakeholders to  extract high-level insights. We address this by integrating LLMs to compress and organize rules into semantically meaningful representations at both global and local levels.

For rule compression, we first transform Boolean expressions into natural language descriptions using LLMs. These unstructured text enables sentence embedding models (e.g., all-mpnet-base-v2) to generate latent rule representation. We then cluster these embeddings to group structurally different but semantically similar rules to identify common patterns across high-performing rules. This reveals latent ``rule families'' that capture shared decision logic and can be interpreted as representative \textit{user personas} or \textit{behavioral segments}.

At the global level, LLM generates population-level explanations by summarizing these clusters via key patterns and feature combinations that characterize each group of rules. This provides a higher-level view of model behavior beyond individual Boolean expressions.

At the local level, LLM produces instance-specific explanations by conditioning on both the Boolean rule and an individual sample and how they contribute to the prediction. The full prompt template is provided in Appendix~\ref{appendix:individual_prompt}.

\section{Experiments}
We implement \textsc{BoolXllm} using the GPT‑5.2 across all stages, including feature selection, threshold recommendation, and rule interpretation. Low temperature is used for feature and threshold generation to ensure near-deterministic behavior, while slightly higher variability is allowed for interpretation to improve linguistic quality. Invalid outputs (e.g., nonexistent features) are filtered via a consistency check.

\subsection{Dataset \& Comparisons}
We evaluate on the UCI Bank Marketing dataset~\cite{uci_bank_marketing}, which includes 20 demographic, financial, and interaction features, e.g., \textit{age}, \textit{balance}, \textit{duration}, \textit{previous outcome} with the goal of predicting product subscription.
We report accuracy, precision, recall, F1, and balanced accuracy. Interpretability is assessed via proxy measures: rule complexity (number of literals), threshold readability, and qualitative clarity. \textsc{BoolXai} with all features and standard discretization serves as the baseline.

\subsection{Results on Feature Selection}
We compare \textsc{BoolXai} with all features against \textsc{BoolXllm} with LLM-guided feature selection. The LLM proposes to filter irrelevant variables and prioritizes semantically meaningful ones.
In our experiments with 20 features, 4 features (e.g., \textit{day\_of\_week}) are discarded yielding a 20\% reduction in the feature space while retaining conceptually meaningful variables for marketing campaign.
Even with reduced features \textsc{BoolXllm} achieves a balanced accuracy at 86\% comparable to \textsc{BoolXai}'s performance using all features. 

\subsection{Results on Threshold Recommendations}
We compare LLM-generated thresholds with standard binning. 
Unlike arbitrary discretization, we observe alignment between LLM-generated thresholds and domain semantics.
For example, \textsc{BoolXai} with default binning using the threshold of \texttt{duration $>$ 393} for the last\_contact\_duration feature while \textsc{BoolXllm} recommends \texttt{duration $>$ 400} without any loss in predictive power.

\subsection{Rule Diversity and Semantic Clustering}
Table~\ref{table:rules} in the Appendix shows representative high-performing rule. As can be seen, arbitrarily different rules achieve similar performance which we group using embedding-based clustering. This supports our approach in Section 3.3 where compression identifies coherent rule families and summarizes them as interpretable personas.

\subsection{Results on Interpretation}
Let us compare a \textsc{BoolXai} rule \texttt{Or(duration $>$ 550, pdays $\leq$ 100, month $=$ mar)} with LLM explanations for a test sample (\texttt{age $=$ 53, duration $=$ 141, pdays $=$ 6, poutcome = success})




\smallskip
\noindent\textbf{Rule Compression and Global Interpretation:} 
``\textit{A long call often indicates engagement. If a customer stays on the phone for 9+ minutes, they are probably interested, asking questions, or at least not rejecting the offer outright. If the contact is recent, customer may still remember the previous interaction, or the bank may have identified them as a warm lead. A seasonal campaign performed well in March which might indicate customers had more liquidity, e.g., tax refunds in some countries, and bank’s marketing strategy was more effective that month, e.g., active campaign."}


\smallskip
\noindent\textbf{Local Interpretation:}
``\textit{This customer will subscribe because recent contact is a strong signal that the customer is still engaged or reachable.}''

\section{Related Work}
\label{appendix:related_work}
Recent works use LLMs to improve interpretability, mainly post-hoc for language descriptions of black‑box models
~\cite{wei2022cot,huang2023llm_xai}. These methods often lack fidelity because the reasoning is not tied to the model's actual decision process and risk hallucinations~\cite{jacovi2020towards}. Others incorporate LLMs into ML pipelines for feature selection and structured reasoning~\cite{Li2024exploring,pang2024generating} that typically remain loosely connected to semantics. In contrast, \textsc{BoolXllm} grounds all LLM outputs in explicit Boolean rules learned by \textsc{BoolXai}, using LLM only to translate symbolic logic. This yields faithful, consistent explanations while extending LLM involvement across feature selection, thresholding, and rule interpretation.

\bibliographystyle{named}
\bibliography{aaai25}

\clearpage
\appendix
\section{Feature Selection Prompt}
\label{appendix:feature_selection_prompt}

\begin{lstlisting}[basicstyle=\ttfamily\small, breaklines=true]
Your task is to select the most relevant features from a given dataset to improve the performance of your model.

Dataset Description: <DATASET_URL>

Downstream Model: <MODEL_TYPE>

Please follow these steps to perform feature selection:
- Study the context and column definition of each feature in the dataset.
- Analyze the correlation between features and the target variable.
- Use business sense to determine which features are most relevant to the problem at hand.
- Decide which features to keep and which to discard based on your analysis.
- Provide a rationale for your feature selection decisions.

Strictly evaluate the features provided in the dataset and do not include any additional features that are not present in the dataset.

The output should be in json format, containing the following keys:
- "selected_features": ...
- "discarded_features": ...
- "rationale": ...

Return strictly the json output without any additional text or explanations.
\end{lstlisting}

\section{Threshold Recommendation Prompt}
\label{appendix:threshold_prompt}

\begin{lstlisting}[basicstyle=\ttfamily\small, breaklines=true]
BoolXAI is a rule classifier that can be used as an interpretable model for classification tasks. It works by selecting a subset of features based on a specified threshold. The threshold determines which features are considered important and should be included in the model.

Your task is to suggest threshold values for numerical features in the dataset to improve the performance of the BoolXAI model.

Numerical features are: <numerical_features_list>

Please follow these steps to provide your recommendations:
- Analyze the distribution of each numerical feature in the dataset.
- Use business sense to determine appropriate threshold values for each numerical feature based on their distribution and correlation with the target variable.
- Provide a rationale for your threshold recommendations, explaining how they can improve the performance of the BoolXAI model.

The output should be in json format, containing the following keys:
- "threshold_recommendations": A dictionary where the keys are the names of the numerical features and the values are the recommended threshold values for those features in list format.
- "rationale": A detailed explanation of why you recommended those specific threshold values for each numerical feature, including insights from the distribution analysis, correlation with the target variable, and business sense considerations.
\end{lstlisting}

\section{Rule Explanation Prompt}
\label{appendix:individual_prompt}

\begin{lstlisting}[basicstyle=\ttfamily\small, breaklines=true]
The rules learnt by boolxai are as follows: <rules>

Classify the following candidate based on the above rules: <candidate_data>

Please provide a detailed explanation of the classification decision, including which rules were applied and how they influenced the final classification outcome.

The output should be in json format, containing the following keys:
- "classification": The predicted class for the candidate based on the applied rules.
- "applied_rules": A list of the rules that were applied in the classification process.
- "explanation": Provide global (population-level) explanation using the rules and summarize: What general patterns and insights lead to positive vs negative outcomes; How the key features drive decisions; High-level profiles or behavioral segments implied by the rules. Provide local (instance-level) explanation using the rules and summarize: How the applied rules influence the classification decision, including any interactions between the rules and how they contributed to the final outcome. This should include an analysis of the candidate's data in relation to the conditions specified in the rules. Use business sense, and consider the implications of the rules' conditions to provide a comprehensive explanation.
\end{lstlisting}

\section{Representative Rules and Clusters}
\label{appendix:rules_table}
\newpage
\begin{sidewaystable*}
\centering
\scriptsize
\begin{tabular}{clccccc}
\hline
\textbf{Cluster} & \textbf{Rules learned by \textsc{BoolXai} and compressed and interpreted by \textsc{BoolXllm}} & \textbf{Balanced Accuracy} & \textbf{Accuracy} & \textbf{Precision} & \textbf{Recall} & \textbf{F1} \\
\hline

0 & \texttt{Or(month=mar, duration>550, pdays<100, nr.employed<=5076, month=apr)}
  & 0.863 & 0.821 & 0.379 & 0.917 & 0.536 \\

0 & \texttt{\makecell[l]{Or(And(duration>400, $\sim$pdays<=3), Choose1($\sim$month=mar, $\sim$month=apr),\\
~~~Choose1($\sim$nr.employed<=5076, $\sim$euribor3m>5))}}
  & 0.859 & 0.780 & 0.335 & 0.961 & 0.497 \\

1 & \texttt{\makecell[l]{Or(And(duration>400, Or($\sim$month=nov, $\sim$job=services)), \\
~~~AtMost1($\sim$nr.employed<=5050, AtMost1(default=no, euribor3m<=4), month=may))}}
  & 0.8603 & 0.7868 & 0.3408 & 0.9552 & 0.5024 \\

1 & \texttt{Or(And(euribor3m<=2, $\sim$month=may), job=student, duration>550, month=dec)}
  & 0.8552 & 0.8192 & 0.3744 & 0.9017 & 0.5291 \\

2 & \texttt{AtMost1(And($\sim$duration>550, $\sim$month=apr, $\sim$month=mar), $\sim$nr.employed<=5099)}
  & 0.8613 & 0.8240 & 0.3820 & 0.9095 & 0.5380 \\

2 & \texttt{Choose1(And($\sim$nr.employed<=5076, $\sim$month=oct, $\sim$month=mar, $\sim$duration>500), $\sim$day\_of\_week=OTHER)}
  & 0.8594 & 0.8506 & 0.4212 & 0.8707 & 0.5677 \\

\hline
\multicolumn{7}{l}{\textit{Cluster 0: High engagement or recently contacted customers (pdays, duration)}} \\
\multicolumn{7}{l}{\textit{Cluster 1: Niche segments driven by specific profiles or economic signals (jobs, balances)}} \\
\multicolumn{7}{l}{\textit{Cluster 2: Exclusion-based rules capturing unlikely responders (maintaining adequate recall due to relaxed logical constraints (e.g., \texttt{AtMost1}, \texttt{Choose1}))}} \\
\hline
\end{tabular}
\caption{Example Boolean rules learned by \textsc{BoolXai} and compressed and interpreted by \textsc{BoolXllm} grouped by semantic clusters. Multiple different rules achieve similar performance. Clustering reveals similar patterns captured by each group.}
\label{table:rules}
\end{sidewaystable*}

\end{document}